\documentclass[letterpaper, 10 pt, conference]{ieeeconf}
\IEEEoverridecommandlockouts    %
\usepackage[T1]{fontenc}
\usepackage{newtxtext,newtxmath}

\usepackage{graphics}
\usepackage{graphicx}
\usepackage{amsmath}
\usepackage{dsfont}
\usepackage{booktabs}
\usepackage{color}
\usepackage{hyperref}
\usepackage{cite}
\usepackage{xspace}
\usepackage{pifont}
\usepackage[capitalise]{cleveref}

\usepackage[font=footnotesize]{caption}
\usepackage[font=footnotesize]{subcaption}

\newcommand*{\cnum}[1]{\ding{\numexpr171+#1\relax}} %

\title{\LARGE \bf Removing the Compute Bottleneck for Event-based Visual Place Recognition with Spiking Neural Networks}

\author{Somayeh Hussaini, Tobias Fischer, and Michael Milford
\thanks{The authors are with the QUT Centre for Robotics, Queensland University of Technology, Brisbane, QLD 4000, Australia (e-mail: {\tt\footnotesize s.hussaini@qut.edu.au}).}%
}

\begin{document}
\thispagestyle{empty}
\pagestyle{empty}

\maketitle

\begin{abstract}

Neuromorphic sensing and computing promise low latency and high energy efficiency for robotics, yet most event-based visual place recognition (VPR) pipelines reintroduce dense, frame-like computation downstream of an inherently sparse sensor, forfeiting the sensor's latency advantage. In this paper, we address the problem of resolving an event-based VPR query in tens of microseconds on a commodity CPU. We train a two-layer spiking neural network (SNN) with additive spike-timing-dependent plasticity (STDP), which drives the learned weight matrix towards a strongly bimodal distribution, amenable to aggressive magnitude pruning. At inference, our pipeline exploits this sparsity by reading only the pruned weights corresponding to the query's active pixels. As surviving weights share a near-uniform magnitude, each neuron's score reduces to an integer count of matching synapses instead of a weighted sum, lowering per-query latency. This computation is small enough to remain resident in a CPU's fast on-chip cache. On three event-based VPR benchmarks, our sparse SNN exceeds the recall of LENS, an existing event-based SNN, while reducing per-query latency by $5.5-6.2\times$ to tens of microseconds. It matches the recall of a dense variant of our network and of the sum-of-absolute-differences (SAD) baseline, while reading $10\times$ fewer weights than the dense variant. We further characterise how recall and latency scale with the number of active pixels and reference places. These results suggest that the sparsity of both event-based input and learned weights can already be exploited for low-latency inference on conventional processors.

\end{abstract}

\section{Introduction}
\label{sec:introduction}

Visual place recognition (VPR) allows a robot to recognise where it is by matching its current query frame against a database of previously visited places, and is a key component of robotic navigation~\cite{lowry2016tro, masone2021acce, zhang2021visual}.
In many robotic applications operating under tight size, weight, and power (SWaP) budgets, for example, search-and-rescue robots operating in disaster environments, navigation typically cannot rely on an on-board GPU. 
Under these constraints, the place recognition prediction should add negligible delay to the overall navigation pipeline, leaving sufficient processing time for other necessary on-board tasks, such as planning and control.

Neuromorphic computing, which represents and processes information as sparse, asynchronous events for high energy efficiency and low latency~\cite{schuman2017survey}, is naturally suited to both these constraints and to VPR itself. 
Event cameras report only per-pixel brightness changes, giving microsecond temporal resolution, high dynamic range, and very low power~\cite{gallego2020event}, while spiking neural networks (SNNs) compute with sparse binary spikes rather than dense activations and can run efficiently on neuromorphic hardware~\cite{davies2021advancing}.
However, event-based VPR pipelines~\cite{fischer2020event, lee2021eventvlad, kong2022event} typically reconstruct this sparse stream into dense frames and process them with dense computation downstream, giving up the sparsity these components depend on for efficiency.

\begin{figure}[t]
\centering
\includegraphics[width=\columnwidth]{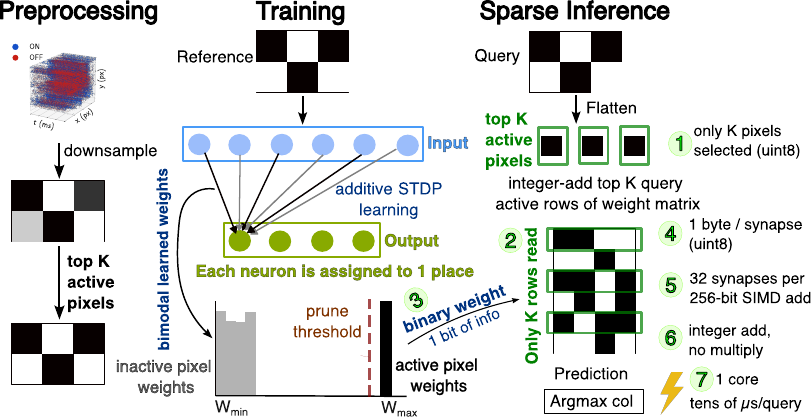}
\caption{\textbf{Method overview.}
\emph{Preprocessing:} event windows are accumulated into frames, downsampled, and reduced to their top-$K$ active pixels.
\emph{Training:} additive STDP with winner-take-all mechanism assigns one neuron per reference place and makes the weights bimodal. A threshold between the modes prunes the weight matrix and leaves \cnum{3} a binary weight matrix. 
\emph{Sparse inference:} a query's \cnum{1} $K$ active pixels select \cnum{2} only $K$ rows of the weight matrix. With each weight stored as \cnum{4} one \texttt{uint8}, \cnum{5} a single 256-bit SIMD instruction sums 32 weights using \cnum{6} integer additions. 
The argmax gives the prediction \cnum{7} in tens of microseconds on one CPU core.}
\label{fig:method}
\vspace{-0.2cm}
\end{figure}

We address the problem of performing event-based VPR with low per-query latency on a commodity CPU, which is difficult for two reasons.
First, avoiding a GPU suits SWaP-constrained platforms, as it reduces on-board components and eliminates host-to-device transfer. However, it also removes the massive parallelism that existing pipelines rely on.
Second, and less obviously, matching a single query is a tiny computation. 
Modern CPUs can sustain hundreds of billions of floating-point operations per second, but only on large, batched workloads that keep them continuously busy. 
A robot navigating online cannot buffer queries and process them in bulk. It must resolve each one as it arrives. 
At this scale, the per-query SNN arithmetic is negligible, and latency is instead dominated by the surrounding overhead (library and kernel-launch calls, memory allocation, and GPU data transfer) and by how many bytes must move from memory per query. 
Therefore, designing a low-latency event-based VPR system on a commodity CPU means matching the algorithm to what the hardware and its libraries can deliver for a single query, not simply reducing FLOP counts.

Existing event-based VPR pipelines~\cite{fischer2020event, lee2021eventvlad, kong2022event, fischer2022many}, including SNN-based methods such as LENS~\cite{hines2025compact}, typically process dense, frame-based representations on a GPU using general-purpose frameworks such as PyTorch~\cite{paszke2019pytorch} and spiking extensions such as snnTorch~\cite{eshraghian2023training}. At the scale of a single VPR query, this puts most of the latency in framework and data-transfer overhead rather than actual computation, so the low-latency potential of event cameras and SNNs is not yet fully realised.

The main contribution of this paper is an event-based SNN for visual place recognition whose training procedure results in a prunable weight matrix, and whose inference pipeline harnesses this sparsity, together with the sparsity of the event input, to achieve lower per-query latency than existing event-based SNN methods for VPR on a commodity CPU (\Cref{fig:method}). This brings total per-query latency into the tens-of-microseconds regime without a GPU or specialised neuromorphic hardware.

In particular, we make the following contributions:
\begin{itemize}

    \item We introduce an event-based SNN for VPR trained with additive STDP under winner-take-all competition, driving the learned weight matrix to a strongly bimodal distribution. A single global magnitude threshold, chosen to best separate each neuron's active-pixel weights from its inactive ones, then prunes the majority of the matrix.

    \item At inference, our pipeline reads only the weights of the query's active pixels and replaces each weight's multiply-accumulate with an integer-add over a compact one-byte-per-weight mask, as surviving weights are all close to a single magnitude. This brings total per-query latency into the tens-of-microseconds regime.

    \item We benchmark our system against LENS, an existing event-based SNN, and a sum-of-absolute-differences (SAD) baseline, under matched preprocessing and on three event-based VPR datasets (QCR-Event-VPR, Brisbane-Event-VPR, and NSAVP). We further analyse the factors contributing to this latency reduction. We also characterise how recall and latency scale with the number of active pixels per query, and how latency scales with the number of reference places learned.
\end{itemize}

\section{Related Work}

We review prior work on low-latency visual place recognition, covering conventional and event-based approaches (\Cref{RW:low_latency_vpr}), and then the efficient sparse and low-precision computation that our method builds on (\Cref{RW:efficient_computation}).

\subsection{Low-latency visual place recognition}
\label{RW:low_latency_vpr}

Visual place recognition (VPR) recognises a previously visited place by matching a query against a database of reference places, with challenges in appearance and viewpoint variations~\cite{lowry2016tro, masone2021acce, zhang2021visual}.

Prior works have investigated optimising the per-query compute latency for image-based VPR mainly by reducing the network and its descriptors. 
\cite{ferrarini2022binary} binarised weights and activations to a single bit, and~\cite{ferrarini2022highly} combined it with depthwise-separable convolutions to further reduce compute. 
\cite{grainge2024design} explored the impact of quantisation bit-width on recall and latency. 
Among bio-inspired approaches, \cite{hussaini2022spiking} proposed a spiking network for VPR, and VPRTempo~\cite{hines2024vprtempo} used time-to-first-spike encoding to reach over 50 Hz on CPU.

Event cameras have been used for VPR due to their microsecond temporal resolution, high dynamic range, and low power consumption. 
Most approaches accumulate events into a frame over a fixed window or event count, and match it with a conventional dense descriptor. 
\cite{fischer2020event} combined descriptors from several window lengths, 
EventVLAD~\cite{lee2021eventvlad} denoised reconstructed edges before matching, 
and Event-VPR~\cite{kong2022event} trained an end-to-end network to directly process the event stream. 
\cite{fischer2022many} tracked event counts from a small, fixed subset of high-variance pixels for an entire traverse, avoiding dense reconstruction altogether.

Spiking networks have also been incorporated with event data. 
LENS~\cite{hines2025compact} deployed an end-to-end neuromorphic pipeline on a mobile robot, 
and SpikeVPR~\cite{keime2026event} learned compact place descriptors and an augmentation method for robustness to speed and temporal variation. 
Flash~\cite{ramanathan2025prepare} matched directly on active pixel regions using bitwise operations, reaching sub-millisecond sensor-to-decision latency.
We differ in that we reduce the compute needed to process an accumulated event frame through a trained spiking network, whereas Flash matches directly on raw events to reach sub-millisecond sensor-to-decision latency.

\subsection{Efficient sparse and low-precision computation on commodity processors}
\label{RW:efficient_computation}

On a CPU, arithmetic is cheap but fetching data from main memory is slow, so processors keep recently used data in \emph{cache}, which is a small, fast on-chip memory organised in levels (L1-L3) that trade speed for capacity~\cite{hennessy2025computer}.
Therefore, a matrix-vector product's runtime on a CPU is governed less by its FLOP count than by how many bytes it must move, whether that data fit in cache, and how well its memory access pattern vectorises, i.e., lets a single single-instruction, multiple-data (SIMD) instruction process many elements at once.
We build on three lines of work that accelerate such computations, which are sparse representations, weight pruning, and low-precision arithmetic.
Sparse matrix multiplication is a well-studied memory-bandwidth-bound kernel~\cite{gao2023systematic}, and sparse structures have been used to improve the energy efficiency of deep networks~\cite{han2016eie} and Transformers~\cite{zhuang2023survey}.
A parallel line of work prunes network weights to unstructured sparsity, skipping pruned entries to reduce inference latency and memory footprint. \cite{han2015learning} learned important connections and pruned the rest, and~\cite{han2015deep} added trained quantisation with Huffman coding. These methods typically target large matrices in a bandwidth-bound regime, or throughput-oriented batch inference, where the fixed cost of an irregular memory access is amortised over many outputs that reuse it.

A third line of work uses low-precision or binary weights so a single instruction covers more elements, turning multiplication into cheaper bitwise or integer operations.~\cite{hubara2016binarized} trained binarised networks and \cite{hubara2018quantized} trained quantised networks, both reducing memory consumption while still computing a cheap per-element product.
At small problem sizes, too few outputs exist to amortise memory-access cost as large-batch matrix-multiply libraries do, so small-batch kernels require different blocking and parallelisation strategies~\cite{georganas2018anatomy}.
Existing low-latency VPR works, including event-based and spiking ones, typically evaluate a dense representation, while sparse and low-precision computation is usually studied for large matrices or batched throughput. Our approach combines these ideas at the scale of a single VPR query using binary, cache-resident weights.

\section{Methodology}

Our pipeline has three stages: each event window is reduced to a sparse binary frame (\Cref{method:event_based}), a two-layer SNN is trained on the reference frames so that each output neuron represents one place (\Cref{method:training}), and queries are matched by a sparse inference pipeline (\Cref{method:inference}).
Our preprocessing and training build on established components.
Like~\cite{fischer2022many}, we represent each event window with a sparse subset of pixels, but we reselect the $K$ most active pixels for every frame instead of fixing one high-variance subset for the entire traverse.
We train the network with additive STDP~\cite{song2000competitive}, combined with winner-take-all competition, adaptive thresholds, and weight normalisation, which were introduced for digit recognition~\cite{diehl2015unsupervised} and later adapted to VPR~\cite{hussaini2022spiking}.
Similarly, sparsity, low-precision weights, and cache locality are well-established ways to accelerate matrix computation~\cite{han2016eie, hubara2016binarized, georganas2018anatomy}, and low-precision networks have been applied to image-based VPR~\cite{ferrarini2022binary, grainge2024design}.
To our knowledge, these techniques have not been combined to reduce event-based VPR latency.

\subsection{Event-based processing}
\label{method:event_based}

We accumulate events, each a tuple of pixel, timestamp, and polarity, over fixed-duration windows onto a downsampled grid of $N_\text{in}$ cells.
A high-contrast scene can produce millions of events per window, making preprocessing cost unbounded, so we cap each window at $E_{\max}$ events, kept as $B$ equal-length runs of consecutive events spaced evenly across the window.
We sample contiguous runs as opposed to striding over individual events because the CPU copies data into cache in 64-byte blocks (cache lines), each holding thirty-two 16-bit event indices. Contiguous runs use nearly every byte of each fetched line, whereas strided sampling fetches a new line for nearly every event it keeps.
Finally, we rank each cell $i$ by its event count $n_i$ and keep the $K$ highest-count cells,
producing a binary frame $\mathbf{x} \in \{0,1\}^{N_\text{in}}$ that is sparse by construction:
\begin{equation}
    x_i = \mathds{1} [i \in \operatorname*{top-K}_{i' \in \{1,\dots,N_\text{in}\}} n_{i'} ],
    \label{eq:event_count}
\end{equation}
where $\mathds{1}[\cdot]$ is the indicator function.

\subsection{Spiking Network training}
\label{method:training}

\subsubsection{Architecture and neuron model}
\label{method:architecture}

Our two-layer network has an input layer of $N_\text{in}$ pixels and an output layer of $N_\text{out} = R$ leaky integrate-and-fire (LIF) neurons, one for each of the $R$ reference places. 
The layers are fully connected by a learnable weight matrix $\mathbf{W} \in \mathbb{R}^{N_\text{in} \times N_\text{out}}$, where $W_{ij}$ is the weight from pixel $i$ to neuron $j$. 
During training, each binary frame $\mathbf{x}$ (\Cref{eq:event_count}) is presented as a Poisson spike train of $T$ steps, in which pixel $i$ emits an input spike $s^\text{pre}_i(t) \in \{0,1\}$ with probability $\rho x_i$ at each step $t \in \{1,\dots,T\}$.
Each neuron $j$ receives the input current $I_j(t) = \sum_i s^\text{pre}_i(t) W_{ij}$, and its membrane potential $V_j$ evolves as:
\begin{equation}
    \tau \frac{dV_j}{dt} = -(V_j - v_\text{rest}) + r I_j,
    \label{eq:lif}
\end{equation}
where $\tau$ is the membrane time constant, $v_\text{rest}$ the resting potential, and $r$ the membrane resistance. We integrate \Cref{eq:lif} using forward Euler with a time step $\Delta t$. When $V_j$ exceeds its threshold (\Cref{method:competition}), neuron $j$ emits a spike and is held at the reset potential $v_\text{reset}$ for a refractory period $\tau_\text{ref}$.

\subsubsection{Additive STDP learning}
We train $\mathbf{W}$ with spike-timing-dependent plasticity (STDP), which updates each weight using only the spikes of its pixel and neuron, with no global error signal.
Each pixel $i$ and neuron $j$ keeps an activity trace, $A^\text{pre}_i$ and $A^\text{post}_j$, which increases by 1 on each of its spikes and decays exponentially with time constants $\tau_\text{pre}$ and $\tau_\text{post}$, respectively.
Following~\cite{song2000competitive}, at each step we apply an additive update followed by a hard clip:
\begin{equation}
    W_{ij} \leftarrow \operatorname{clip} ( W_{ij} + A_{+}A^\text{pre}_i s^\text{post}_j - A_{-}A^\text{post}_j s^\text{pre}_i,\; W_{\min}, W_{\max}),
    \label{eq:stdp}
\end{equation}
where $s^\text{post}_j \in \{0,1\}$ indicates a spike of neuron $j$ at the current step, $s^\text{pre}_i$ is the input spike, $A_{+}$ and $A_{-}$ are the potentiation and depression amplitudes, and $\operatorname{clip}(\cdot)$ limits its argument to $[W_{\min}, W_{\max}]$.
A pixel spiking before its neuron strengthens their weight, and spiking after weakens it. Setting $A_{+} > A_{-}$ lets strengthening dominate for pixels that repeatedly fire with their neuron.
To set the weight scale, we define $w^* = v_\text{th}\tau / (r \Delta t K)$, the per-pixel weight at which a single Euler step of \Cref{eq:lif} from rest reaches $v_\text{th}$ for a frame with $K$ active pixels.
We initialise $W_{ij} \sim \mathcal{U}(0, w^*)$ and set $[W_{\min}, W_{\max}] = [0, 2w^*]$, i.e.~excitatory synapses only.

\subsubsection{Competition, homeostasis and supervision}
\label{method:competition}

At each step, among neurons whose potential exceeds their effective threshold $v_\text{th} + \theta_j + o_j$, only the one with the largest margin spikes, and all membrane potentials are reset. 
The homeostatic term $\theta_j$ increases by $\theta_{+}$ per spike and decays by a factor of $\gamma_\theta$ after each presentation, preventing any one neuron from winning every place. 
We assign neuron $p$ to place $p$ before training, and, while place $p$ is presented, set $o_p = -b$, lowering its threshold by $b$ ($o_j = 0$ otherwise). 
This soft bias breaks close contests between neurons sharing active pixels, preventing overlapping places from converging onto one neuron.

\subsubsection{Weight normalisation}
\label{method:normalisation}

After each presentation, each neuron's incoming weights are rescaled to a fixed $L_1$ budget $W_\text{sum}$, their expected sum at initialisation:
\begin{equation}
    W_{ij} \leftarrow \operatorname{clip} ( W_{ij} \frac{W_\text{sum}}{\sum_{i'} W_{i'j}}, W_{\min}, W_{\max} ).
    \label{eq:w_norm}
\end{equation}
As STDP raises the sum above $W_\text{sum}$, rescaling lowers all of a neuron's weights, but subsequent potentiation restores only those of its place's active pixels.
Because each STDP update is a fixed step, these weights saturate towards $W_{\max}$ while the rest decay towards $W_{\min}$, yielding the bimodal distribution that enables pruning (\Cref{method:prune}).

\subsection{Inference pipeline}
\label{method:inference}

At inference, each query frame $\mathbf{x}$ (\Cref{eq:event_count}) is presented once and drives the input current directly, without Poisson sampling. After a single Euler step of \Cref{eq:lif} from rest, each neuron's membrane potential is proportional to that current, so we rank places directly by it:
\begin{equation}
    c_j = \sum_{i=1}^{N_\text{in}} x_i W_{ij}, \qquad \hat{p} = \operatorname*{argmax}_j c_j
    \label{eq:inf}
    \vspace{-0.1cm}
\end{equation}
where $c_j$ is the score of neuron $j$, and hence of its place, and $\hat{p}$ is the predicted place.

\subsubsection{Pruning}
\label{method:prune}

Before inference, we prune $\mathbf{W}$ using the neuron-to-place assignment (\Cref{method:competition}), since neuron $j$ should be driven only by the pixels active in its place's reference frame.
This splits all synapses $(i,j)$ into an \emph{active} set $\mathcal{S}^{+}$, where pixel $i$ is on in neuron $j$'s reference frame, and an \emph{inactive} set $\mathcal{S}^{-}$, where it is off.
We choose the single global threshold $\eta$ that best separates the two sets by maximising Youden's $J$ statistic~\cite{youden1950index}:
\begin{equation}
    \eta = \operatorname*{argmax}_{\eta'} [ f_{\mathcal{S}^{+}}(\eta') - f_{\mathcal{S}^{-}}(\eta') ],
    \label{eq:prune_threshold}
\end{equation}
where $f_\mathcal{S}(\eta')$ is the fraction of weights in $\mathcal{S}$ with $W_{ij} \ge \eta'$.
Here, $f_{\mathcal{S}^{+}}$ and $f_{\mathcal{S}^{-}}$ act as the true- and false-positive rates, so $\eta$ keeps as many active weights as possible while discarding as many inactive ones as possible.
Setting every weight below $\eta$ to zero gives the pruned matrix $\tilde{\mathbf{W}}$, with entries $\tilde{W}_{ij} = W_{ij} \mathds{1}[W_{ij} \ge \eta]$.

\subsubsection{Sparse inference}
\label{method:sparse}

Only the $K$ rows of $\tilde{\mathbf{W}}$ indexed by the query's set of active pixels $\mathcal{Q}$ contribute to \Cref{eq:inf}, as $x_i = 0$ elsewhere.
As surviving weights share a near-uniform magnitude (\Cref{method:normalisation}), their magnitudes add little ranking information, so we store only a one-byte-per-synapse binary mask $M_{ij} = \mathds{1}[\tilde{W}_{ij} \neq 0]$.
Each neuron's score then becomes an integer count of its surviving synapses to the query's active pixels:
\begin{equation}
    c_j = \sum_{i \in \mathcal{Q}} M_{ij},
    \label{eq:sparse}
\end{equation}
which gives the same prediction as \Cref{eq:inf} when the surviving weights are equal. \Cref{res:weight_quality} shows that they do saturate at $W_{\max}$ over the operating range we use.
Thus, each query reduces to summing $K$ mask rows, costing $O(K N_\text{out})$ instead of $O(N_\text{in} N_\text{out})$.
Compared with float32 weights, the mask reads $4\times$ fewer bytes per synapse, fits $4\times$ more synapses per SIMD instruction, and replaces multiply-accumulate with integer addition (\Cref{res:latency}).

\section{Experimental Setup}
\label{sec:experimentalsetup}

We describe our implementation details in~\Cref{exp:imp_details}, the baselines in~\Cref{exp:baselines}, the evaluation metrics in~\Cref{exp:evaluation_metrics}, and the datasets in~\Cref{exp:datasets}.

\subsection{Implementation details}
\label{exp:imp_details}

We take all measurements on an Ubuntu machine with a single Intel Core i7-10700K CPU (8 physical cores, 16 threads, 3.8 GHz base / 5.1 GHz turbo, 32 KiB L1d and 256 KiB L2 per core, 16 MB shared L3) with 32 GB RAM and an NVIDIA GeForce RTX 3080 (10 GB).
All preprocessing runs on the CPU. Our sparse SNN and SAD run as single-core Numba-compiled CPU kernels and our dense SNN as an 8-core one, while LENS runs on the GPU via PyTorch, following its published implementation.

Frames are $32 \times 32$ ($N_\text{in} = 1024$) with $K = 100$ active pixels, taken from 100 ms windows on QCR-Event and 1 s windows on Brisbane-Event and NSAVP, keeping both event polarities, with each window capped at $E_{\max} = 20000$ events drawn as $B = 256$ contiguous runs. 
The output layer has one neuron per reference place ($N_\text{out} = R$). 
The LIF neurons use $\Delta t = 10^{-3}$ s, $\tau = 10^{-2}$ s, $\tau_\text{ref} = 5\times10^{-3}$ s, $r = 1$, $v_\text{th} = 0.5$ and $v_\text{rest} = v_\text{reset} = 0$. 
Each training presentation is a Poisson spike train of $T = 125$ steps at $\rho = 0.1$.
We train the network for 30 epochs over the reference set.
STDP uses $\tau_\text{pre} = \tau_\text{post} = 0.02$ s with amplitudes set as fractions of $W_{\max}$, $A_{+} = 0.015\cdot W_{\max}$, $A_{-} = 0.003 \cdot W_{\max}$, so they stay valid as $K$ changes.
Homeostasis uses $\theta_{+} = 0.05\cdot v_\text{th}$ per spike with decay rate of $\gamma_\theta = 0.99$. The supervision bias is $b = 0.5 v_\text{th}$.

\subsection{Baselines}
\label{exp:baselines}

We compare against two external baselines, LENS and SAD, and a dense variant of our pipeline. 
All methods use our preprocessed event frames, so recall and latency are compared under the same inputs.

\subsubsection{LENS}
\label{exp:lens}
\cite{hines2025compact} is our primary event-based SNN baseline, a compact VPR network deployed on both a GPU and SynSense Speck neuromorphic processor~\cite{yao2024spike}. It is the most relevant comparison for our method.
Unlike our two-layer network, LENS adds a feature layer between the pixel input and the per-place output layer.
The input-to-feature weights are learned unsupervised via STDP, and the feature-to-output weights with a supervised delta rule, to allow each neuron to learn a single place.  
It encodes frames with time-to-first-spike coding, so more active pixels spike earlier, and uses LIF neurons.
We use LENS's published network, training and plasticity code unmodified on our preprocessed frames (\Cref{method:event_based}), and run its inference on GPU.

\subsubsection{Sum of Absolute Differences (SAD)}
\label{exp:sad}
\cite{milford2012seqslam} is a simple learning-free VPR method that returns the reference frame with the smallest pixel-wise difference to the query.
We acknowledge that it does not achieve state-of-the-art R@1. However, we include it as a conventional method whose very fast computation makes it a good comparison for compute speed.
For a fair latency comparison, we optimised SAD. We compute its distance as the Hamming distance between bit-packed binary frames using XOR and popcount instructions. This reduces the reference database by $32\times$ relative to float32 and keeps it in fast cache.

\subsubsection{Dense SNN backend}
\label{exp:dense}

To isolate where the sparse backend's speedup comes from, this dense variant of our pipeline computes \Cref{eq:inf} as a matrix-vector multiply over the same pruned matrix $\tilde{\mathbf{W}}$, reading all $N_\text{in} N_\text{out}$ weights per query.
For a fair latency comparison, we optimise it by storing each neuron's weights as a contiguous row for unit-stride vectorised reads, and by splitting neurons into row blocks across all 8 cores so no cross-thread merge is needed.

\begin{figure}
    \begin{subfigure}{\columnwidth}
        \centering
        \includegraphics[width=\textwidth]{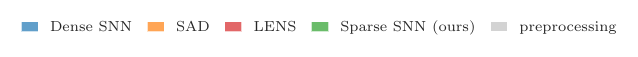}
    \end{subfigure}
    \vspace*{-0.3cm}
    \begin{subfigure}{\columnwidth}
        \centering
        \includegraphics[width=0.9\textwidth, trim= 0cm 0.6cm 0cm 0cm, clip]{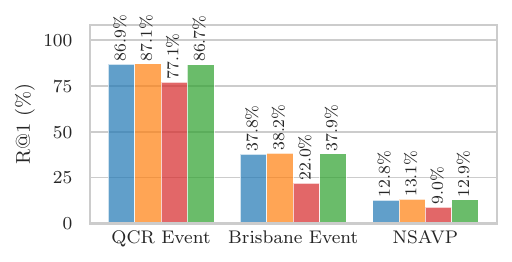}
    \end{subfigure}
    \centering
    \begin{subfigure}{\columnwidth}
        \centering
        \includegraphics[width=0.9\textwidth]{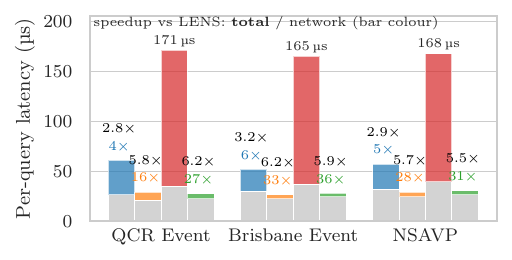}
    \end{subfigure}
    \vspace*{-0.2cm}
    \caption{
    R@1 and latency performance comparisons across all methods and datasets. 
    [top] Top-1 R@1: our sparse SNN exceeds LENS's R@1 and matches dense SNN's and SAD's R@1 across all datasets.
    [bottom] Average per query latency, divided into preprocessing (light gray) and network time (solid colour). Labels give each method's total (black) and network-only (solid color) speedup over LENS. 
    LENS's network stage dominates its latency, whereas our sparse SNN and SAD reduce network time to single digits. Dense SNN has a higher network latency than our sparse SNN but it is still substantially lower than LENS's. 
    }
    \label{fig:main}
    \vspace*{-0.4cm}
\end{figure}

\subsection{Evaluation metrics}
\label{exp:evaluation_metrics}

We report recall at 1 (R@1), the fraction of queries whose top-1 prediction is correct, and the average per-query latency. 
We consider a query as correctly matched if the top-1 reference match is within $\pm 10$ reference frames of the ground-truth match, which is equivalent to $\pm 1$s for the QCR-Event and $\pm 10$s for Brisbane-Event and NSAVP.
To suppress timing outliers, we run each query 10 times, take the median latency, and report the mean over all queries in each dataset.

\subsection{Datasets}
\label{exp:datasets}

We evaluate on three event-based VPR datasets. 
QCR-Event-VPR~\cite{fischer2022many} and Brisbane-Event-VPR~\cite{fischer2020event} were both recorded with a DAVIS346 camera. 
QCR-Event covers a 160 m indoor route traversed 16 times with varying speeds and camera orientations. We use normal2 as the reference and normal3 as the query.
Brisbane-Event follows an 8 km urban route traversed six times at ${\sim} 50 km/h$, with natural stops and varying illumination and weather. We use sunset1 as the reference and sunset2 as the query.
The Novel Sensors for Autonomous Vehicle Perception (NSAVP) dataset~\cite{carmichael2025dataset} used a stereo DVXplorer event camera ($640 \times 480$). 
We use its 8.3-km moderate-density urban route, driven at a mean speed of 9 m/s, with the sunset traverse ($R0\_FS0$) as the reference and the afternoon traverse ($R0\_FA0$) as the query.

\section{Results}
\label{sec:results}

\subsection{Recall and latency performance comparisons}
\label{res:main}

We first compare against LENS, the most relevant baseline. 
Our sparse SNN exceeds LENS's R@1 on all three datasets (86.7\% vs 77.1\% on QCR-Event, 37.9\% vs 22.0\% on Brisbane-Event, 12.9\% vs 9.0\% on NSAVP) while reducing total per-query latency by $5.5 - 6.2\times$ (\Cref{fig:main}).
The gap comes almost entirely from LENS's network stage, where PyTorch's GPU kernel launches, memory allocations, and host-device transfers add a large fixed overhead to every query.
In contrast, our sparse SNN runs as one Numba-compiled kernel call per query on a single CPU core, with call overhead of only a few percent of its latency (\Cref{fig:inference_cost}), and reads only 0.16 MB of weights per query, which fits in L2 cache. This reduces its network latency to single-digit microseconds.

We also compare against two baselines with distinct roles: the dense SNN, which uses the same weights and inputs, and so isolates the latency effect of sparse inference (\Cref{exp:dense}), and SAD, a fast learning-free conventional VPR method.
The R@1 of our sparse SNN closely matches that of the dense SNN and SAD across all three datasets, while its average per-query latency is lower than the dense SNN's and very similar to SAD's.
The dense SNN also achieves a total latency that is $2.8 - 3.2\times$ lower than LENS's across all datasets.

The R@1 and latency of our sparse SNN are very similar to SAD's. 
This is because, after pruning, each neuron's mask closely matches its place's reference frame (\Cref{res:weight_quality}), so its score counts the active pixels shared by the query and that reference. As every frame has exactly $K$ active pixels, SAD's Hamming distance equals $2K$ minus twice this overlap, resulting in near-identical rankings.
However, our sparse SNN recovers these reference frames through locally-driven STDP learning, with no global error signal, whereas SAD is a learning-free method that uses bitwise operations.
Thus, reaching a competitive decision boundary from purely local plasticity is a non-trivial validation of our training and pruning pipeline.
Moreover, our sparse SNN learns through local STDP and scores places by synaptic integration, the two primitives that neuromorphic hardware such as Loihi~\cite{davies2021advancing} is built to support.
Hence, deploying it on such hardware with on-chip plasticity, where new places could be learned directly from the event stream, is a promising future direction, which learning-free SAD does not offer.

\subsection{Weight bimodality and pruning quality}
\label{res:weight_quality}

\begin{figure}[t]
    \begin{subfigure}{\columnwidth}
        \centering
        \includegraphics[width=\textwidth]{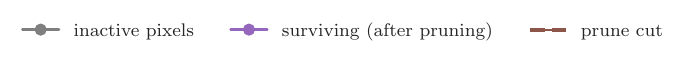}
    \end{subfigure}
    \vspace{-0.1cm}
    \begin{subfigure}{\columnwidth}
        \centering
        \includegraphics[width=0.9\linewidth]{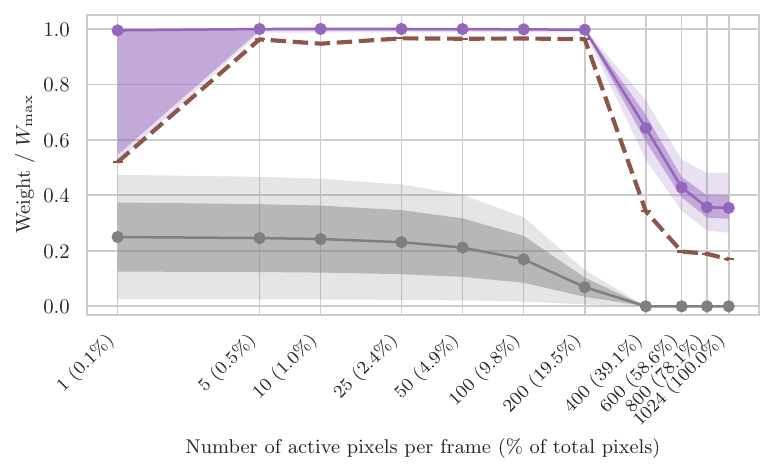}
    \end{subfigure}
    \vspace*{-0.2cm}
    \caption{
    Learned weights across varying number of active pixels. 
    Solid lines show the median line of the inactive pixel and surviving weights, with shaded bands of the same colour spanning the 25–75th (darker) and 5–95th (lighter) percentiles. 
    The dashed line is the pruning threshold $\eta$ (\Cref{eq:prune_threshold}). 
    From 5 to 200 px, surviving weights are saturated at $W_\text{max}$ and clearly separated from the inactive weights. With few active pixels, the inactive weights remain close to their uniform initialisation, $U(0, 0.5 W_\text{max})$. From 400 px, the active pixels need more than each neuron's fixed weight budget. Then, normalisation drives the inactive weights to zero, while the surviving weights fall below $W_\text{max}$ and become more variable (median 0.36 $W_\text{max}$ at 1024 px).
    }
    \label{fig:weight_quality}
    \vspace*{-0.3cm}
\end{figure}

\Cref{fig:weight_quality} shows how the learned weights change with the number of active pixels $K$.
The pattern follows from the weight budget in \Cref{eq:w_norm}. The incoming weights of every neuron sum to $W_\text{sum} = N_\text{in} w^*/2 = (N_\text{in}/4)W_{\max}$ for any $K$, which is $256W_{\max}$ for our $32 \times 32$ frames. 
This budget determines the distribution of the active and inactive weights.
When $K < N_\text{in}/4$, the budget is large enough for every active weight to saturate at $W_{\max}$. 
From 5 to 200 active pixels, the surviving weights are all at $W_{\max}$ with almost no variability, and the pruning threshold falls just below them. 
This shows that the learned weights are strongly bimodal in this region, and replacing them with a binary mask (\Cref{eq:sparse}) preserves what was learned. 
The leftover budget remains on the inactive weights, which is why they stay close to their initial values when $K$ is small and shrink as $K$ grows (median $0.25\cdot W_{\max}$ at 1 px, $0.07\cdot W_{\max}$ at 200 px).
When $K > N_\text{in}/4$, saturating every active weight would exceed the budget. Then, normalisation drives all inactive weights to zero and distributes the budget over the active weights, so their mean falls to about $(N_\text{in}/4K)W_{\max}$. 
This bound causes the median surviving weight to fall to $0.64\cdot W_{\max}$ at 400 px and $0.36\cdot W_{\max}$ at 800 and 1024 px.
Hence, active and inactive weights remain separable at every $K$, but for $K > N_\text{in}/4$ the surviving weights no longer share a common ceiling.
Then, their magnitudes carry ranking information that the binary mask (\Cref{eq:sparse}) discards, and as this spread widens, the sparse SNN's R@1 drops sharply at 800 and 1024 active pixels (\Cref{fig:scalability_num_active_px}).

\subsection{Where latency comes from}
\label{res:latency}

\begin{table}[t]
\centering
\caption{
Per-query inference cost of the dense and sparse backends on QCR-Event ($N_\text{in}=1024$, $N_\text{out}=1600$, $K=100$). 
The top panel gives the work each backend must perform, in arithmetic, memory traffic and issued AVX2 instructions (256-bit registers hold 8 float32 or 32 uint8 lanes), which were verified with hardware performance counters.
Every ratio is the dense value divided by the sparse value.
The bottom panel is a breakdown of the measured $6.73\times$ kernel speedup compared to dense SNN into its contributing factors.
}
\label{tab:inference_cost}
\begin{tabular}{lccc}
\toprule
 & Dense & Sparse & Ratio \\
\midrule
\multicolumn{4}{l}{\emph{Per-query cost}} \\
Element operations   & $N_\text{in}N_\text{out}$ & $K N_\text{out}$ & \\
                     & $1.64\times10^{6}$ & $1.60\times10^{5}$ & $10.2\times$ fewer \\
Weight bytes read    & $4N_\text{in}N_\text{out}$ & $K N_\text{out}$ & \\
                     & $6.55$ MB & $0.16$ MB & $41.0\times$ fewer \\
Vector instructions  & $N_\text{in}N_\text{out}/8$ & $K N_\text{out}/32$ & \\
                     & $2.05\times10^{5}$ & $5.00\times10^{3}$ & $41.0\times$ fewer \\
Threads              & 8 & 1 & \\
Kernel latency       & $34.44 \mu$s & $5.12 \mu$s & $6.73\times$ faster \\
\midrule
\multicolumn{4}{l}{\emph{Decomposition of the $6.73\times$}} \\
\multicolumn{3}{l}{Fewer weights visited ($N_\text{in}/K$)} & $10.24\times$ faster \\
\multicolumn{3}{l}{Single-threaded (dense gains $3.209\times$ from 8 cores)} & $3.209\times$ slower \\
\multicolumn{3}{l}{Accumulator in memory, not registers} & $1.228\times$ slower \\
\multicolumn{3}{l}{uint8 mask, 32 SIMD lanes, integer add} & $2.588\times$ faster \\
\multicolumn{3}{l}{\textbf{Net}\quad $10.24 \times 2.588/ (3.209 \times 1.228)$} & $\mathbf{6.73\times}$ \textbf{faster} \\
\bottomrule
\end{tabular}
\vspace*{-0.3cm}
\end{table}

\begin{figure}
    \begin{subfigure}{\columnwidth}
        \centering
        \includegraphics[width=0.8\textwidth]{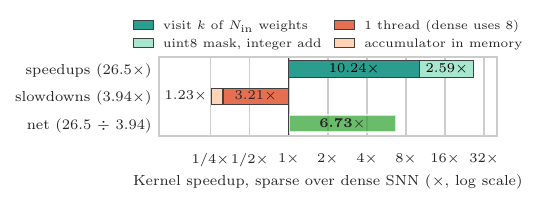}
    \end{subfigure}
    \begin{subfigure}{\columnwidth}
        \centering
        \includegraphics[width=0.8\textwidth]{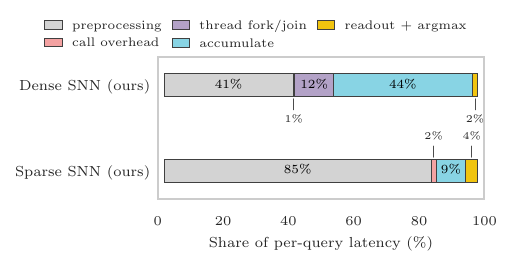}
    \end{subfigure}
    \vspace*{-0.2cm}
    \caption{
    [Top] Breakdown of the measured $6.73\times$ kernel speedup of our sparse SNN over the dense SNN on QCR-Event ($N_\text{in}=1024$, $N_\text{out}=1600$, $k=100$) into the factors of \Cref{tab:inference_cost}.
    [Bottom] Share of total per-query latency attributed to each pipeline stage, for the same setting. The stages are preprocessing, per-call overhead, thread fork/join, weight accumulation, and readout (argmax). 
    Preprocessing takes similar absolute time in both backends, so its larger share for sparse (85\% vs 41\%) reflects the sparse network's lower cost. Thread fork/join appears only for the dense backend, which parallelises across 8 cores. Sparse runs single-threaded. Accumulation dominates the dense's network cost (44\%), whereas the sparse's integer-add accumulation is comparatively small (9\%).
    }
    \label{fig:inference_cost}
    \vspace*{-0.3cm}
\end{figure}

\Cref{tab:inference_cost} and \Cref{fig:inference_cost} details the per-query cost of our sparse backend compared to its dense version on QCR-Event, showing how each design choice in \Cref{method:sparse} contributes to speedup.
The dense backend evaluates the full weight matrix, performing $N_\text{in}N_\text{out}=1.64\times10^{6}$ multiply-accumulate operations, which means reading $4N_\text{in}N_\text{out}=6.55$ MB of float32 weights.
The sparse backend visits only the $K=100$ rows corresponding to the query's active pixels, which is $KN_\text{out}=1.60\times10^{5}$ integer additions across $0.16$ MB of uint8 weights. 
This is $N_\text{in}/K = 10.24\times$ fewer operations and weights read.
Because a 256-bit AVX2 register holds 32 uint8 lanes or 8 float32 lanes, it also means $41\times$ fewer bytes read and vector instructions issued.

The two backends' weight data also sit in different levels of the cache hierarchy, where smaller levels are faster.
The sparse backend's per-query weight data ($0.16$ MB) fits in a single core's private 256-KiB L2 cache, whereas the dense backend's weight matrix ($6.55$ MB) must be read per query from the larger but slower 16-MB L3 cache, which is why we parallelise the dense backend across all 8 cores.

The measured $6.73\times$ speedup is smaller than the $10.24\times$ reduction in weights read, because the sparse's memory layout, which enables skipping rows, incurs a less efficient computation speed. 
First, the sparse backend iterates over input pixels, so all $N_\text{out}$ values accumulate simultaneously in an L1-resident array rather than in a CPU register, which is a storage slot inside the processor core, faster than L1.
As per our measurements, sparse performs $0.129$ memory writes per accumulated element, against $0.001$ for dense, whose scalar accumulator stays in a register, which makes sparse $1.23\times$ slower.

Second, parallelising the sparse backend would require per-thread accumulators and a merge pass, which cost more than they save at this size.
Thus, it runs single-threaded and gives up the dense backend's $3.21\times$ multithreading gain.
Against these combined $3.94\times$ slowdowns, uint8 integer addition is $2.59\times$ faster per element than dense's float32 multiply-accumulate, as it fits $4\times$ more SIMD lanes.
Overall, sparse is $10.24 \times 2.59 / 3.94 = 6.73\times$ faster than dense, reducing per-query network latency from $34.44 \mu$s to $5.12 \mu$s.
\Cref{fig:inference_cost} shows that preprocessing dominates the sparse backend's latency (85\%). 
Hence, sparse's lower latency compared to dense is due to the arithmetic work it eliminates outweighing the parallelism it gives up.

\subsection{Scalability}

\begin{figure}
    \begin{subfigure}{\columnwidth}
        \centering
        \includegraphics[width=\textwidth]{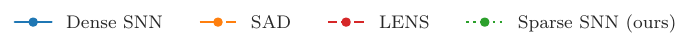}
    \end{subfigure}
    \begin{subfigure}{\columnwidth}
        \centering
        \includegraphics[width=0.85\textwidth]{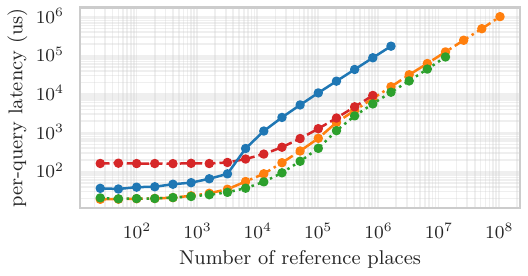}
    \end{subfigure}
    \vspace*{-0.35cm}
    \caption{
    Scalability of per-query latency with the number of reference places for our sparse SNN, dense SNN, SAD and LENS.
    Our sparse SNN scales similar to SAD throughout, flat and lowest below $\sim10^3$ places, and increasing roughly linearly while remaining the fastest. 
    Dense SNN's latency increases moderately until $\sim10^3$ places and grows the fastest beyond that. 
    LENS stays flat at $130 - 140 \mu s$ while its fixed GPU overhead dominates its negligible per query compute, then increases linearly once that compute becomes significant. 
    }
    \vspace*{-0.5cm}
    \label{fig:scalability_num_places}
\end{figure}

Here, we discuss the scalability as the number of reference places and the number of active pixels increases.

\subsubsection{On increasing number of reference places}
\Cref{fig:scalability_num_places} reports the average per query latency as the number of reference places increases.
It only isolates how latency scales with database size, and not R@1 because evaluating R@1 at that large scale would require reference data at that scale and correspondingly long training times. For this experiment only, we used randomised weight matrices, measuring latency across 200 query images per database size.
Our sparse SNN and SAD perform very similarly by staying flat and lowest below $\sim10^3$ places, then increase roughly linearly with the number of reference places, remaining the fastest pair of methods. 
Dense's cost is $O(N_\text{in} N_\text{out})$, so its latency grows fastest, and it is measured only up to ${\sim}1.5\times10^6$ places, past which the weight matrix becomes too large.
LENS's latency is flat at $130-140\mu s$ for small databases, because at this problem size its per query compute is negligible relative to the fixed overhead of launching PyTorch kernels on the GPU and allocating memory. 
Its latency starts to rise once the number of places is large enough that its compute cost becomes larger than the fixed overhead.

\subsubsection{On increasing the number of active pixels}

\begin{figure}
    \begin{subfigure}{\columnwidth}
        \centering
        \includegraphics[width=0.9\textwidth]{images/sweep_legend.pdf} %
    \end{subfigure}
    \vspace*{-0.1cm}
    \begin{subfigure}{\columnwidth}
        \centering
        \includegraphics[width=0.9\textwidth, trim= 0cm 2.4cm 0cm 0cm, clip]{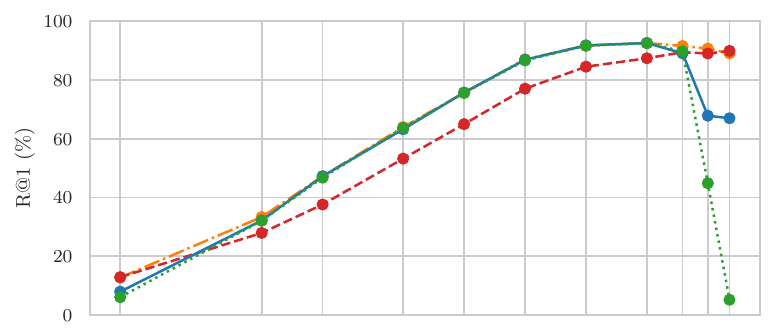}
    \end{subfigure}
    \centering
    \begin{subfigure}{\columnwidth}
        \centering
        \includegraphics[width=0.9\textwidth]{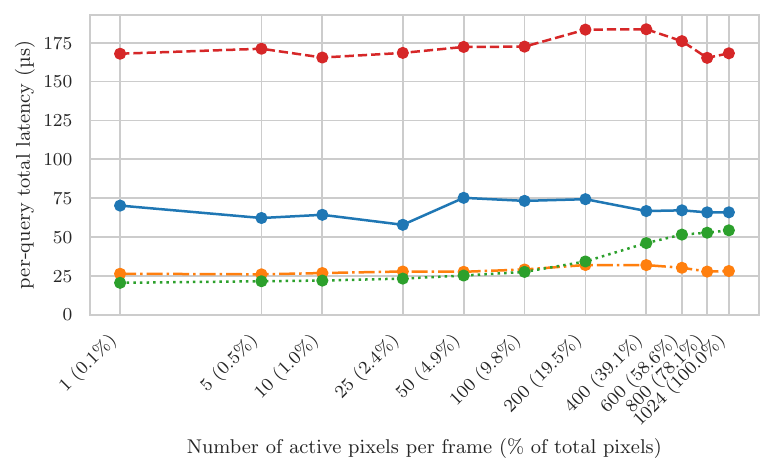}
    \end{subfigure}
    \caption{
    Scalability experiment on how increasing the number of active pixels per frame impacts the R@1 and the per-query latency of our sparse SNN, dense SNN, SAD and LENS.  
    R@1 increases with active pixel count and plateaus at 100–200 pixels (9.8–19.5\% of the image). We chose 100 for balanced accuracy and sparsity. Beyond this, SAD and LENS stay stable, but our dense and sparse models degrade at high pixel counts as denser inputs become harder to discriminate, impacting sparse more as it collapses all active weights to a single value. Latency-wise, sparse tracks SAD up to $\sim$200 pixels before rising with pixel count, while dense stays flat throughout due to always processing the full frame regardless of sparsity.
    }
    \vspace*{-0.35cm}
    \label{fig:scalability_num_active_px}
\end{figure}

\Cref{fig:scalability_num_active_px} shows how R@1 and per-query latency scale with the number of active pixels per frame.
R@1 of our sparse SNN increases with the active pixel count, as it does for other methods, with LENS trailing slightly. 
The performance of all methods then plateaus at 100-200 active pixels (9.8\%-19.5\% of the total pixels in an image). 
Then, SAD and LENS remain fairly flat up to 1024 active pixels, 
whereas dense SNN and our sparse SNN show a moderate to significant R@1 decline at 800 and 1024 active pixels (78.1\% and 100\% of the total pixels).
The sparse SNN is affected more severely, as it ignores weight magnitudes and treats all surviving synapses as equal, causing its R@1 to drop sharply (44.9\% at 800 active pixels and 5.2\% at 1024).
We use 100 active pixels for our main results, as this lies within the R@1 plateau region.

In terms of latency, our sparse SNN follows SAD closely and stays flat up to about 200 active pixels, after which it grows gradually as its computational cost scales with the active pixel counts.
Even with this increase, its latency at 1024 active pixels is slightly lower than dense's. 
The dense SNN's latency remains roughly constant throughout, because it processes the full input frame and weight matrix, and hence is not impacted by the active pixel count.
LENS's latency also stays roughly constant, as its fixed PyTorch and GPU overhead dominates the per-query compute at every pixel count, and stays much higher than that of the other methods.

\section{Discussion and Conclusion}

In this paper, we presented an event-based SNN for VPR whose additive STDP training drives the learned weights towards a strongly bimodal distribution, enabling aggressive pruning.
At inference, this bimodality allows us to binarise the learned weights, while the sparsity of the event input reduces each neuron's weighted sum to an integer count over a one-byte-per-synapse mask that remains resident in a single core's L2 cache.
Across three event-based VPR datasets, our sparse SNN exceeded the R@1 of LENS, an existing GPU-based event SNN, while reducing per-query latency by $5.5-6.2\times$ to tens of microseconds on a single CPU core.
Our latency analysis (\Cref{res:latency}) suggests that, at the scale of a single query, fitting the computation within the CPU cache hierarchy matters more than minimising FLOPs.

Our approach has three main limitations.
First, recall degrades once $K$ substantially exceeds $N_\text{in}/4$, and drops sharply at 800 and 1024 active pixels, as surviving weights lose their common magnitude and places increasingly share active pixels. Future work could suppress pixels active across many places, and retain the most discriminative ones so that integer counting remains informative at higher $K$.
Second, our reported latency excludes the 100 ms to 1 s event accumulation window, so end-to-end sensor-to-decision latency remains bounded by the window duration. Asynchronous per-event updates could remove this bottleneck.
Finally, deployment on neuromorphic hardware such as Loihi~\cite{davies2021advancing} or Speck~\cite{yao2024spike} could realise the corresponding energy-efficiency gains.

{\small
\bibliographystyle{ieee_fullname}
\bibliography{references}
}

\end{document}